\documentclass{bmvc2k}

\usepackage{bm}
\usepackage{cleveref}
\usepackage{amsfonts}
\usepackage{booktabs}
\usepackage{colortbl}
\usepackage{indentfirst}

\newtheorem{intuition}{Intuition}
\newtheorem{formalism}{Formalism}

\title{Cycle Consistency in Video Object-Centric Learning}

\addauthor{Rongzhen Zhao}{rongzhen.zhao@aalto.fi}{1}
\addauthor{Zhiyuan Li}{zhiyuan.li@aalto.fi}{1}
\addauthor{Ruonan Wei}{ruonan2765@gmail.com}{2}
\addauthor{Juho Kannala}{juho.kannala@aalto.fi}{3,4}
\addauthor{Joni Pajarinen}{joni.pajarinen@aalto.fi}{1}

\addinstitution{
Department of Electrical Engineering and Automation,\\
Aalto University,\\
Espoo, Finland
}
\addinstitution{
School of Artificial Intelligence and Automation,\\
Huazhong University of Science and Technology,\\
Wuhan, China
}
\addinstitution{
Department of Computer Science,\\
Aalto University,\\
Espoo, Finland
}
\addinstitution{
Center for Machine Vision and Signal Analysis,\\
University of Oulu,\\
Oulu, Finland
}

\runninghead{Zhao, Li, Wei, Kannala, Pajarinen}{CC + OCL}

\begin{document}

\maketitle

\begin{abstract}
Self-supervised video Object-Centric Learning (OCL) aims to discover distinct objects and associate them across time, whereas self-supervised Multi-Object Tracking (MOT) focuses on associating pre-defined object detections or segmentations.
Although well-established in MOT, Cycle Consistency (CC) cannot naively or explicitly apply to the latent slot space of OCL.
Unlike the deterministic and ideal object representations in MOT, OCL slots are inherently stochastic and ambiguous due to non-unique scene decompositions. Enforcing explicit cycle consistency (ECC) on slots imposes rigid mean seeking. This severely penalizes the model for exploring alternative but equally valid decompositions, thereby driving towards feature collapse.
To resolve this dilemma, we propose \textit{Implicit Cycle Consistency (ICC)}, which shifts the cycle-consistency constraint from the restrictive slot space to the continuous reconstruction manifold, encouraging slots to reach a soft consensus on collectively interpreting the visual scene rather than forcing rigid point-to-point feature alignment.
Extensive experiments on complex video OCL benchmarks demonstrate that ICC avoids feature collapse and outperforms ECC baselines.
Our source code is provided as the supplement.
\end{abstract}

\section{Introduction}
\label{sect:introduction}

Self-supervised video representation learning has advanced along two representative directions: Multi-Object Tracking (MOT)~\citep{wang2019udt} and Object-Centric Learning (OCL)~\citep{singh2022steve}. 
Self-supervised MOT associates object identities across frames without trajectory labels given ideal object detections or segmentations. 
Whereas, OCL decomposes each video frame into objects (and background) and associate them temporally to represent the visual scene with minimal information loss.
Understanding their connection can support the improvement of video OCL for visual scene representation and understanding~\citep{wu2022slotformer}.

Although MOT has been studied extensively, its insights have rarely been utilized for improving video OCL.
A cornerstone of MOT is \textit{Cycle Consistency} (CC), which enforces agreement between forward and backward object associations~\citep{meng2023contrast}: an object's trajectory tracked from frame $t$ to $t+k$ and back to $t$ should return to the original object, or their forward-backward trajectories should overlap.
Intuitively, this utilizes the video's inherent temporal coherence: if an object can be tracked reversibly with negligible drift, the model learns robust associations directly from raw video.
Incorporating this principle into video OCL suggests a promising route to unify object discovery and temporal association.

\begin{figure}[]
\small
\centering
\includegraphics[width=0.5\linewidth]{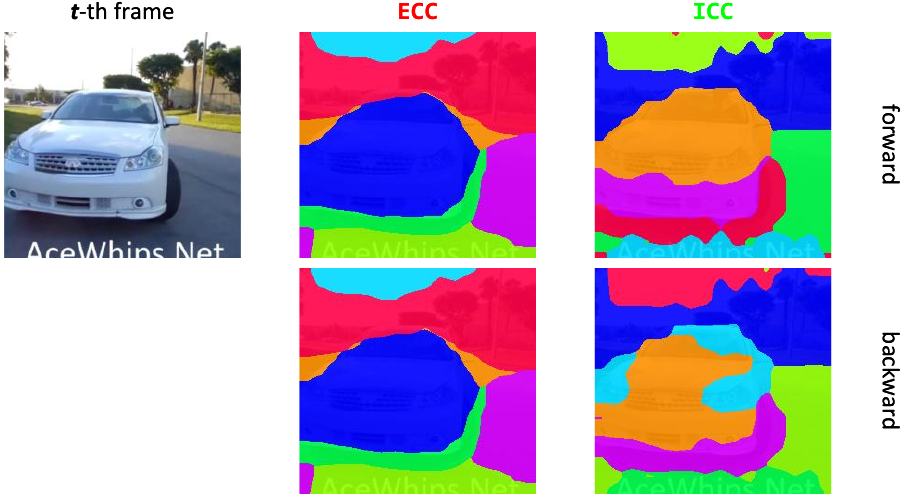}
\caption{
Decomposition Divergence. We visualize the attention masks of frame $t$ from Forward and Backward streams. (\textit{middle}) Explicit Cycle Consistency (\texttt{\textcolor{red}{ECC}}) forces alignment, causing conflict and blurriness. (\textit{right}) Implicit Cycle Consistency (\texttt{\textcolor{green}{ICC}}) allows the Forward stream to clearly segment the car, land, tree and sky, while decomposing the car into different sets of the car's parts. Gaussian-smoothed for better presentation.
}
\label{fig:diverge}
\end{figure}

However, we demonstrate that naively applying explicit CC (ECC) to video OCL is fundamentally ill-posed. 
We identify a core conflict: unlike MOT's ideal object representations, OCL's slot representations are inherently \textit{stochastic} and \textit{ambiguous}. 
The self-supervised decomposition of a visual scene can have different possibilities. When modeling a car, different decompositions may partition it into body + wheels, or alternatively into windshield + hood + the remaining. So the forward and backward streams often converge to distinct, yet equally valid, decompositions \cite{fan2024adaslot}.
Consequently, enforcing \textbf{hard alignment} in the \textbf{slot space} (latent space) penalizes the model for exploring these valid variations. 
As shown in \Cref{fig:diverge} middle and \Cref{sect:cc_explicit}, this drives representations toward a collapsed average, smoothing out discriminative features essential for object discovery.

To address this, we propose \textbf{Implicit CC} (ICC) for video OCL.
We align the forward and backward streams on the \textbf{reconstruction manifold} (observation space), enforcing a \textbf{soft consensus}.
As shown in \Cref{fig:diverge} right and \Cref{sect:cc_implicit}, while allowed to diverge in the latent space to accommodate stochasticity and ambiguity, forward-backward slots must reach a consensus on explaining the visual scene, i.e., the reconstruction.
This leverages the temporal self-supervision of CC without suffering from feature collapse.


In this work we make the following contributions:
(\textit{c1}) We articulate a fundamental conflict when adapting cycle consistency from MOT to video OCL, showing that rigid latent alignment suppresses the slots' capacity to handle scene-decomposition ambiguities.
(\textit{c2}) We enforce temporal consistency implicitly on the reconstruction manifold, allowing slots to maintain optimization flexibility while capturing robust temporal correlations across frames.
(\textit{c3}) ICC improves object discovery on complex video datasets, outperforming explicit alignment baselines while demonstrating strong resistance to feature degradation.

\section{Related Work}
\label{sect:related_work}

We discuss related work organized into three key directions, omitting the attributive ``self-supervised'' for brevity.

\subsection{Image Object-Centric Learning}
\label{sect:image_ocl}

Image OCL serves as the spatial foundation for video OCL.
Most existing methods adopt an encode–aggregate–decode architecture \citep{locatello2020slotattent,zhao2025vvo}.
The encoder usually employs some Vision Foundation Models (VFMs) for feature extraction.
The aggregator is typically based on slot attention \citep{locatello2020slotattent}, which aggregates VFM features into mutually competing slots to obtain object-level representations, i.e., slots; the attention maps of these slots can be used for object segmentation, i.e., object discovery.
The decoder reconstructs the input from slots, providing self-supervision that encourages each slot to capture as much information as possible.

Recent advances can be categorized by the affected modules.
Methods for improving aggregation include \cite{jia2023boqsa,biza2023isa,zhao2025smoothsa}. Methods for improving decoding include \cite{wu2023slotdiffuz,kakogeorgiou2024spot,zhao2025dias}.
Methods for improving reconstruction include \cite{singh2021slate,seitzer2023dinosaur,zhao2025vvo}.

\subsection{Video Object-Centric Learning}
\label{sect:video_ocl}

Video OCL is a temporal extension of image OCL,  performing image OCL on each video frame while connecting frames recurrently via a transition module~\cite{singh2022steve}.
Namely, video OCL not only decomposes each frame into objects but also associate them across time.

Recent advances can be categorized by the affected modules.
Methods improving reconstruction include:
SAVi and SAVi++ \cite{kipf2021savi,elsayed2022savipp} using flow and depth for weakly-supervised object separation signal;
VideoSAUR \citep{zadaianchuk2024videosaur} predicting patch movement for temporal consistency.
Methods improving transitioning include:
SlotContrast \cite{manasyan2025slotcontrast} introducing slot contrastive loss for temporal consistency;
RandSF.Q \cite{zhao2025randsfq} predicting next queries from a random slot-feature pair with relative time information for implicit transition dynamics modeling.

Despite these advances, existing Video OCL methods typically rely on forward-only stream, lacking verifications on forward-backward consistency.

\subsection{Multi-Object Tracking}
\label{sect:mot}

To learning how to associate objects through time without trajectory labels, i.e., self-supervised MOT, some form of consistency has to be utilized.
Cross-view consistency \cite{bastani2021crossview} requires invariance across view perturbations.
Path consistency \cite{lu2024path} enforces stability across variable temporal strides.
Graph-based methods like \cite{segu2024walker} utilize temporal appearance graphs to maintain coherence. 
Temporal consistency \cite{erregue2025yolo11jde} maintains similar appearance in a tracking of one object.
We specifically focus on cycle-consistency (CC) \cite{zhao2023ocmot, meng2023contrast}, where an object trajectory in forward and backward streams should overlap.

In whichever case, explicit consistency regularization is enforced on object representations directly.
This is effective because object representations are extracted from ideal detections or segmentations.
However, this does not hold in video OCL, where naively applying explicit CC is actually harmful.

\section{Proposed Method}
\label{sect:proposed_method}

In this section, we present our framework for integrating cycle consistency into unsupervised video Object-Centric Learning (OCL). We first outline the video OCL formulation based on the most recent state-of-the-art methods, RandSF.Q \cite{zhao2025randsfq} and SmoothSA \cite{zhao2025smoothsa}, where our method is built upon. We then analyze the limitations of applying explicit cycle constraints (common in MOT) to the OCL setting. Finally, we introduce our \textit{Implicit Bi-directional Consensus} mechanism, which leverages stochastic slot dynamics to enforce temporal consistency without hindering object discovery.

\subsection{Preliminary: Video Object-Centric Learning}
\label{sect:p_video_ocl}

Given a video clip of $T$ frames, we extract visual features $\{ \bm{F}_t \in \mathbb{R}^{h \times w \times c} \} _ {t=1} ^ {T}$ using a pre-trained encoder, as formalized in prior work \cite{zhao2025vvo}. The goal of video OCL is to map these features to a set of $S$ slot vectors $\bm{S}_t \in \mathbb{R}^{S \times c}$ for each time step $t$, representing objects and background in the $t$-th video frame.

Denote the typical video OCL process as \textbf{forward stream} $\mathcal{T}_\text{fw}$, which operates as below:
\begin{subequations}
\label{eq:fw}
\begin{align}
\label{eq:initaliz}
& \textit{\textcolor{gray}{initialization}}
& \bm{Q}_1 &= \bm{\phi}_\text{n} (\bm{C})
& t = 1
\\
\label{eq:transit}
& \textit{\textcolor{gray}{transition}}
& \bm{Q}_t &= \bm{\phi}_\text{r} (\bm{S}_{t-1}, \bm{F}_t) + \bm{\epsilon}
& t > 1
\\
\label{eq:aggregat}
& \textit{\textcolor{gray}{aggregation}}
& \bm{S}_t, \bm{A}_t &= \bm{\phi}_\text{a} (\bm{Q}_t, \bm{F}_t)
\\
\label{eq:decode}
& \textit{\textcolor{gray}{decoding}}
& \hat{\bm{F}}_t &= \bm{\phi}_\text{d} (\bm{S}_t)
\end{align}
\end{subequations}
\textit{Initialization}: if $t=1$, initializer $\bm{\phi}_\text{n}$ transforms clue $\bm{C}$, e.g., learned Gaussian samplings \cite{locatello2020slotattent} or object bounding boxes \cite{kipf2021savi}, into slot queries $\bm{Q}_1 \in \mathbb{R} ^ {S \times c}$.
\textit{Transition}: if $t>1$, transitioner $\bm{\phi}_\text{r}$ recurrently transforms previous frame's slots $\bm{S}_{t-1}$ into next queries $\bm{Q}_t$, with training stochasticity $\bm{\epsilon}$ utilized in the most recent state-of-the-art, RandSF.Q \cite{zhao2025randsfq} and SmoothSA \cite{zhao2025smoothsa}.
\textit{Aggregation}: aggregator $\bm{\phi}_\text{a}$, a Slot Attention module or its variants, iteratively aggregates information in feature $\bm{F}_t$ into queries $\bm{Q}_t$, producing slots $\bm{S}_t$, along with byproduct attention maps $\bm{A}_t \in \mathbb{R} ^ {S \times h \times w}$, which can be binarized as object segmentation masks.
\textit{Decoding}: decoder $\bm{\phi}_\text{d}$ decodes slots $\bm{S}_t$ into feature reconstruction $\hat{\bm{F}}_t \in \mathbb{R} ^ {h \times w \times c}$.

The self-supervised is achieved by minimizing the reconstruction error of all time steps:
\begin{equation}
\label{eq:fw_loss}
\mathcal{L}_\text{recon} = \sum _ {t=1} ^ T \| \hat{\bm{F}}_t - \bm{F}_t \|^2
\end{equation}
Note that there might be some auxiliary losses \cite{zadaianchuk2024videosaur,manasyan2025slotcontrast}, which are ignored for brevity.

\begin{figure*}
\centering
\small
\includegraphics[width=\textwidth]{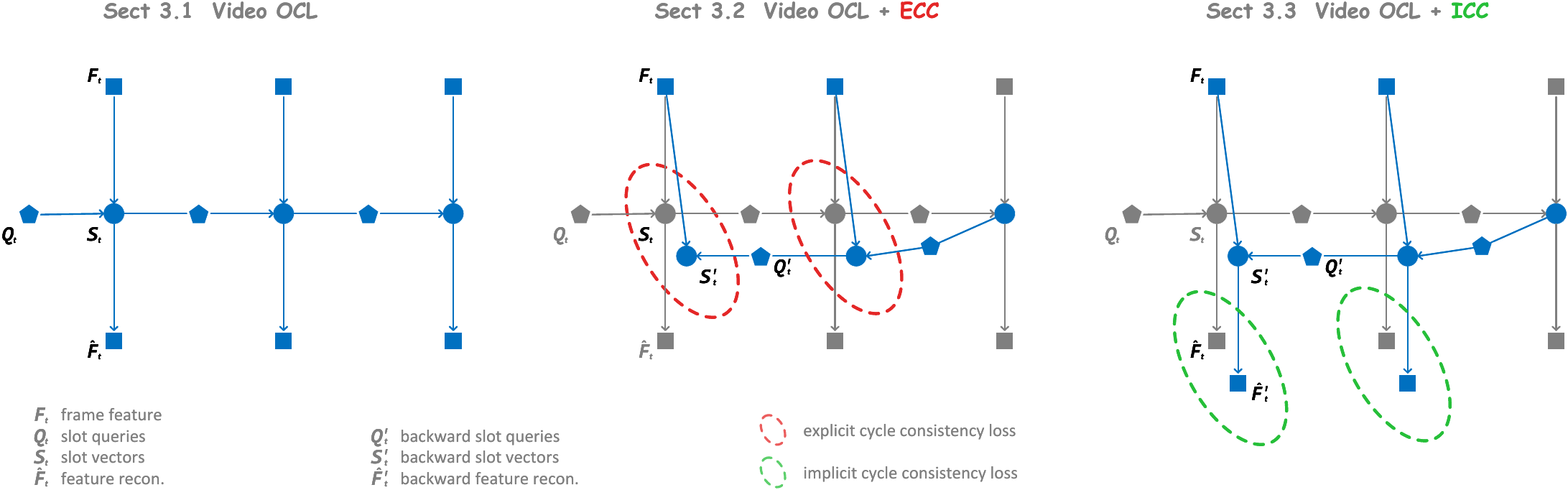}
\caption{
Cycle Consistency in video OCL.
(\textit{left}) Baseline video OCL with forward-only stream.
(\textit{middle}) Explicit Cycle Consistency (\texttt{\textcolor{red}{ECC}}) applies a loss directly on forward-backward slots, forcing hard latent alignment. We demonstrate this is ill-posed and leads to feature collapse due to decomposition ambiguity.
(\textit{right}) Implicit Cycle Consistency (\texttt{\textcolor{green}{ICC}}) applies the loss on forward-backward feature reconstruction. By aligning on the observation manifold rather than the latent space, ICC enforces explanation power while preserving latent representation diversity.
}
\label{fig:solution}
\end{figure*}

\subsection{Explicit Cycle Consistency}
\label{sect:cc_explicit}

A naive intuition is to borrow technique Cycle Consistency (CC) from self-supervised Multiple Object Tracking (MOT) \cite{lu2024path,meng2023contrast}, to enforce \textit{Cycle Consistency}, which crucial in associating objects across time without external supervision:
\begin{subequations}
\label{eq:bw_explicit}
\begin{align}
\label{eq:bw_initializ_e}
& \textit{\textcolor{gray}{initialization}}
& \bm{Q}'_{T-1} &= \bm{\phi}'_\text{r} (\bm{S}_{T}) + \bm{\epsilon}
& t = T - 1
\\
\label{eq:bw_transit_e}
& \textit{\textcolor{gray}{transition}}
& \bm{Q}'_t &= \bm{\phi}'_\text{r} (\bm{S}'_{t+1}, \bm{F}_t) + \bm{\epsilon}
& t < T - 1
\\
\label{eq:bw_aggregat_e}
& \textit{\textcolor{gray}{aggregation}}
& \bm{S}'_t, \bm{A}'_t &= \bm{\phi}_\text{a} (\bm{Q}'_t, \bm{F}'_t)
\end{align}
\end{subequations}
\textit{Initialization}: if $t=T-1$, backward transitioner $\bm{\phi}'_\text{r}$ transforms the last forward slots $\bm{S}_T$ into the first backward queries $\bm{Q}'_{T-1}$.
\textit{Transition}: if $t < T-1$, transitioner $\bm{\phi}'_\text{r}$ recurrently transforms previous backward slots $\bm{S}'_{t+1}$ into the next backward queries $\bm{Q}'_t$.
\textit{Aggregation}: aggregator $\bm{\phi}_\text{a}$ iteratively aggregates information in feature $\bm{F}_t$ into backward queries $\bm{Q}_t$, producing backward slots $\bm{S}'_t$, along with byproduct backward attention maps $\bm{A}'_t$.

Through CC, an object's forward and backward associations through time should have overlapped trajectories.
Mathematically, this implies an explicit regularization loss:
\begin{equation}
\label{eq:bw_loss_explicit}
\mathcal{L}_\text{ECC} = \sum _ {t=1} ^ {T-1} \| \bm{S}'_t - \bm{S}_t \|^2
\end{equation}
where the explicit CC loss $\mathcal{L}_\text{ECC}$ is added into the original losses for joint optimization.

\begin{intuition}
\label{proposit1}
(\textbf{Feature Collapse from Decomposition Divergence})
Applying explicit cycle consistency to video OCL is ill-posed. The unsupervised decomposition of a scene is inherently ambiguous, i.e., existing multiple valid ways to segment complex objects. Explicit alignment penalizes the model for exploring different decompositions, forcing collapse to averaged representation.
\end{intuition}

\begin{formalism}
In the forward stream $\mathcal{T}_\text{fw}$, let the set of slots $\bm{S}_t$ represent a decomposition hypothesis $\bm{H}_t \in \mathbb{R} ^ {S \times c}$ derived from the feature map $\bm{F}_t$, where each $\bm{H}_{t,s}$ corresponds to a discovered visual entity.
In the backward stream $\mathcal{T}_\text{bw}$, starting from $T$ and propagating back, the accumulated stochastic noise $\bm{\epsilon}$ leads the model to sample a different decomposition mode $\bm{H}'_t$.
Crucially, given the self-supervision, $\bm{H}_t$ and $\bm{H}'_t$ are likely to differ not just in permutation, but in content, e.g., $\bm{H}_{t,s}$ captures ``rider + bike rear wheel'' while $\bm{H}'_{t,s}$ captures ``rider + bike front wheel''.
The explicit consistency loss imposes:
\begin{equation}
\label{eq:proposit1_loss_cce}
\mathcal{L}_\text{ECC} = \| \bm{S}_t - \bm{S}'_t \|^2 \approx
\sum _ s
\| \bm{H}_{t, s} - \bm{H}'_{t, \pi(s)} \|^2
\end{equation}
where $\pi$ is an implicit matching. Since $\bm{H}_t \neq \bm{H}'_t$ due to decomposition divergence, the gradient acts to minimize the distance between two distinct visual concepts:
\begin{equation}
\label{eq:proposit1_grad_cce_s}
\nabla_{\bm{S}_t} \mathcal{L}_\text{ECC} = 2(\bm{S}_t - \bm{S}'_t) \approx 2(\bm{H}_t - \bm{H}'_t)

\end{equation}
This creates a ``mean-seeking'' force:
Instead of refining the segmentation, this force pulls the slot representations towards the average of the diverging hypotheses $\mathbb{E}[\bm{H}_{t,:}]$, smoothing out discriminative features and leading to the observed representation collapse (blurriness or background absorption).
\end{formalism}

\subsection{Implicit Cycle Consistency}
\label{sect:cc_implicit}

To effectively utilize temporal information without the drawbacks of explicit alignment, we introduce a symmetric \textbf{backward stream} $\mathcal{T}_{\text{bw}}$ and an \textbf{implicit consensus} objective.

Subsequent the forward stream, we instantiate a backward stream that processes the video frames inversely from $t=T$ down to $1$:
\begin{subequations}
\label{eq:bw_implicit}
\begin{align}
\label{eq:bw_initializ_i}
& \textit{\textcolor{gray}{initialization}}
& \bm{Q}'_{T-1} &= \bm{\phi}'_\text{r} (\bm{S}_{T}) + \bm{\epsilon}
& t = T - 1
\\
\label{eq:bw_transit_i}
& \textit{\textcolor{gray}{transition}}
& \bm{Q}'_t &= \bm{\phi}'_\text{r} (\bm{S}'_{t+1}, \bm{F}_t) + \bm{\epsilon}
& t < T - 1
\\
\label{eq:bw_aggregat_i}
& \textit{\textcolor{gray}{aggregation}}
& \bm{S}'_t, \bm{A}'_t &= \bm{\phi}_\text{a} (\bm{Q}'_t, \bm{F}'_t)
\\
\label{eq:bw_decode_i}
& \textit{\textcolor{gray}{decoding}}
& \hat{\bm{F}}'_t &= \bm{\phi}_\text{d} (\bm{S}'_t)
\end{align}
\end{subequations}
\textit{Initialization}: if $t=T-1$, backward transitioner $\bm{\phi}'_\text{r}$ transforms the last forward slots $\bm{S}_T$ into the first backward queries $\bm{Q}'_{T-1}$.
\textit{Transition}: if $t < T-1$, transitioner $\bm{\phi}'_\text{r}$ recurrently transforms previous backward slots $\bm{S}'_{t+1}$ into the next backward queries $\bm{Q}'_t$.
\textit{Aggregation}: aggregator $\bm{\phi}_\text{a}$ iteratively aggregates information in feature $\bm{F}_t$ into backward queries $\bm{Q}_t$, producing backward slots $\bm{S}'_t$, along with byproduct backward attention maps $\bm{A}'_t$.
\textit{Decoding}: decoder $\bm{\phi}_\text{d}$ decodes backward slots $\bm{S}'_t$ into backward feature reconstruction $\hat{\bm{F}}'$.

Crucially, in the backward stream $\mathcal{T}_\text{bw}$, modules aggregator $\bm{\phi}_\text{a}$ and decoder $\bm{\phi}_\text{d}$ are rigidly shared from the forward stream $\mathcal{T}_\text{fw}$, while transitioner ${\bm\phi}'_\text{r}$ shares most weights with its forward counterpart $\bm{\phi}_\text{r}$ to ensure the model learns time-consistent physical rules, differing only in relative time embeddings \cite{zhao2025randsfq}.

Instead of forcing $\bm{S}_t$ and $\bm{S}'_t$ to be similar, we enforce that both representations must yield the \textit{same} reconstruction of the scene. The implicit consensus is achieved by joint minimizing the forward reconstruction loss \Cref{eq:fw_loss} and the following backward reconstruction loss:
\begin{equation}
\mathcal{L}_\text{ICC} = \sum _ {t=1} ^ {T - 1} \| \hat{\bm{F}}'_t -\bm{F}_t \|^2
\label{eq:bw_loss_implicit}
\end{equation}

\begin{figure*}
\centering
\small
\includegraphics[width=\textwidth]{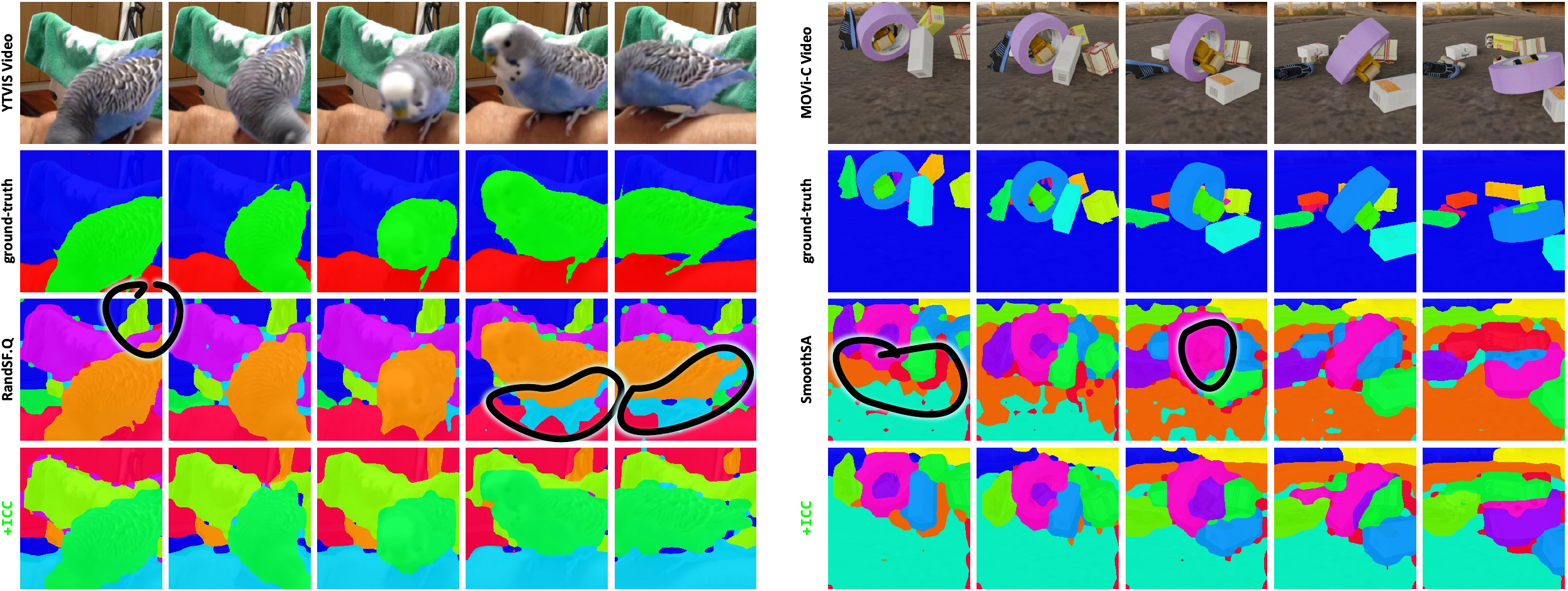}
\caption{
Qualitative results of object discovery on videos. Our \textcolor{green}{ICC} improves basis methods, RandSF.Q~\citep{zhao2025randsfq} and SmoothSA~\citep{zhao2025smoothsa}, consistently. Note that the differences of segmentation colors have no semantic meaning.
}
\label{fig:placeholder}
\end{figure*}

We demonstrate that our proposed implicit objective maximizes the mutual consistency of the bi-directional streams in the observation space, bypassing the permutation problem.

\begin{intuition}
\label{proposit2}
(\textbf{Decomposition Consensus on Reconstruction Manifold})
Unlike explicit alignment, our implicit regularization imposes consistency on the \textit{observation} manifold, allowing slots to diverge in vector space, accommodating \textit{Permutation Drift}, provided they remain functionally equivalent in scene reconstruction.
\end{intuition}

\begin{formalism}
Let us view the video OCL process as a Variational Autoencoder (VAE) framework. We aim to maximize the log-likelihood of video data $\log p(\bm{F}_t)$.
The forward and backward streams approximate the intractable posterior using two distinct variational distributions, $q_1(\bm{S}|\bm{F})$ and $q_2(\bm{S}'|\bm{F})$. The proposed objective, \Cref{eq:fw_loss,eq:bw_loss_implicit}, can be viewed as maximizing the Evidence Lower Bound (ELBO) for both streams simultaneously:
\begin{equation}
\label{eq:proposit2_elbo}
\begin{aligned}
\mathcal{J} 
= \mathbb{E}_{\bm{S} \sim q_1} [\log p_{\bm{\phi}}(\bm{F}|\bm{S})] 
+ \mathbb{E}_{\bm{S}' \sim q_2} [\log p_{\bm{\phi}}(\bm{F}|\bm{S}')] 
- \mathrm{KL}(\dots)
\end{aligned}
\end{equation}
Let 
$\mathcal{M} = \{ \hat{\bm{F}} \mid \exists \bm{S}, \hat{\bm{F}} = \bm{\phi}_\text{d}(\bm{S}) \}$
be the manifold of all possible image reconstructions decodable from the slot space.
The implicit loss requires that:
\begin{equation}
\label{eq:proposit2_require}
\bm{\phi}_\text{d}(\bm{S}_t) \approx \bm{F}_t \quad \text{and} \quad \bm{\phi}_\text{d}(\bm{S}'_t) \approx \bm{F}_t
\end{equation}
Let $\mathcal{S}_{\bm{F}} = \bm{\phi}_{\text{d}}^{-1}(\bm{F}_t)$ be the set of all valid slot configurations (permutations and decompositions) that can reconstruct frame $\bm{F}_t$.
By minimizing the reconstruction error for both streams, we actually enforce:
\begin{equation}
\label{eq:proposit2_enforce}
\bm{S}_t \in \mathcal{S}_{\bm{F}} \quad \text{and} \quad \bm{S}'_t \in \mathcal{S}_{\bm{F}}
\end{equation}
Crucially, since decoder $\phi_{\text{d}}$ is permutation invariant regarding slots \cite{wiedemer2024provable},
the condition $\bm{S}_t, \bm{S}'_t \in \mathcal{S}_{\bm{F}}$ does \textit{not} imply $\bm{S}_t = \bm{S}'_t$.
Instead, it implies consistency in \textit{explanatory power}. The stochastic transitions $\bm{\epsilon}$ allow $\bm{S}_t$ and $\bm{S}'_t$ to explore different regions of $\mathcal{S}_{\bm{F}}$ (the solution space), preventing the model from getting stuck in local minima, while the joint reconstruction objective ensures both paths remain valid explanations of the visual scene.
\end{formalism}

\section{Experiment}
\label{sect:experiment}

We evaluate Implicit Cycle Consistency (ICC) across a hierarchy of visual understanding: unsupervised object discovery (\Cref{sect:objdiscov}) and downstream object recognition (\Cref{sect:objrecogn}).
All experiments are conducted with the same set of three random seeds whenever applicable.
Note that as our ICC is designed as a hyperparameter-free plugin, which can be integrated into state-of-the-art basis methods, there is no need to do any ablation study.

\begin{table*}
\small
\centering
\setlength{\tabcolsep}{2pt}
\newcommand{\tss}[1]{\scalebox{0.8}{\texttt{#1}}}
\newcommand{\cg}[1]{\textcolor{green}{#1}}
\newcommand{\ch}[1]{\textcolor{red}{#1}}
\newcommand{\std}[1]{\scalebox{0.4}{±#1}}

\begin{tabular}{ccccccccccccc}
\toprule
& ARI & ARI\textsubscript{fg} & mBO & mIoU & ARI & ARI\textsubscript{fg} & mBO & mIoU & ARI & ARI\textsubscript{fg} & mBO & mIoU \\
\arrayrulecolor{gray}
\cmidrule(lr){2-5} \cmidrule(lr){6-9} \cmidrule(lr){10-13}
& \multicolumn{4}{c}{MOVi-C {\tiny \#slot=11, conditional}} & \multicolumn{4}{c}{MOVi-E {\tiny \#slot=21, conditional}} & \multicolumn{4}{c}{YTVIS-HQ {\tiny \#slot=7}} \\
\arrayrulecolor{black}
\midrule
VideoSAUR    & 41.9\std{1.1} & 53.3\std{2.1} & 16.1\std{0.4} & 14.8\std{0.4} & 17.4\std{2.5} & 34.6\std{20.7} & 8.3\std{4.9} & 7.5\std{4.3} & 33.8\std{0.7} & 49.2\std{0.5} & 29.9\std{0.4} & 29.7\std{0.4} \\
SlotContrast        & 64.6\std{9.4} & 59.9\std{5.3} & 27.7\std{3.0} & 25.8\std{2.9} & 29.9\std{4.9} & 70.6\std{3.8} & 20.7\std{1.4} & 19.3\std{1.2} & 37.2\std{0.6} & 49.4\std{1.1} & 33.0\std{0.2} & 32.8\std{0.1} \\
\arrayrulecolor{gray}
\cmidrule(lr){1-1} \cmidrule(lr){2-5} \cmidrule(lr){6-9} \cmidrule(lr){10-13}
\arrayrulecolor{black}
RandSF.Q        & 65.4\std{10.7} & 67.4\std{2.1} & 29.2\std{3.8} & 26.8\std{3.7} & 30.5\std{1.2} & 82.1\std{3.1} & 23.0\std{1.2} & 21.6\std{1.4} & 40.1\std{0.4} & 58.0\std{1.0} & 37.6\std{0.4} & 37.2\std{0.4} \\
\quad + \ch{\texttt{ECC}} & \ch{53.8}\std{3.4} & \ch{46.6}\std{3.7} & \ch{20.5}\std{1.1} & \ch{17.6}\std{1.7} & 34.0\std{4.2} & \ch{45.1}\std{2.3} & \ch{13.3}\std{5.6} & \ch{11.8}\std{6.0} & \ch{40.0}\std{1.6} & \ch{57.2}\std{3.5} & \ch{37.0}\std{0.7} & \ch{36.2}\std{1.8} \\
\quad + \cg{\texttt{ICC}} & \cg{73.2}\std{0.7} & 	\cg{67.4}\std{1.0} & 	\cg{32.9}\std{0.4} & 	\cg{30.3}\std{0.3} & 	\cg{41.6}\std{7.5} & 	80.5\std{4.1} & 	\cg{26.3}\std{1.3} & 	\cg{24.8}\std{1.1} & 	\cg{40.6}\std{1.0} & 	\cg{60.1}\std{3.6} & 	\cg{39.2}\std{0.3} & 	\cg{38.9}\std{0.4} \\
\arrayrulecolor{gray}
\cmidrule(lr){1-1} \cmidrule(lr){2-5} \cmidrule(lr){6-9} \cmidrule(lr){10-13}
\arrayrulecolor{black}
SmoothSA    & 50.9\std{1.6} & 69.0\std{0.3} & 31.7\std{0.8} & 30.2\std{0.8} & 36.7\std{0.6} & 73.6\std{0.6} & 28.6\std{0.1} & 27.4\std{0.1} & 42.4\std{0.8} & 63.0\std{3.4} & 38.9\std{0.7} & 38.3\std{0.6} \\
\quad + \ch{\texttt{ECC}} & \ch{44.1}\std{1.6} & 70.4\std{0.1} & \ch{30.6}\std{0.8} & \ch{29.2}\std{0.8} & \ch{35.0}\std{0.8} & \ch{65.2}\std{0.5} & \ch{23.9}\std{0.3} & \ch{22.6}\std{0.2} & \ch{40.2}\std{0.3} & \ch{59.8}\std{1.2} & \ch{38.0}\std{0.3} & \ch{37.2}\std{0.3} \\
\quad + \cg{\texttt{ICC}} & \cg{52.1}\std{2.3} & 	\cg{71.2}\std{0.9} & 	\cg{33.4}\std{0.7} & 	\cg{32.1}\std{0.8} & 	35.4\std{1.2} & 	\cg{74.0}\std{0.3} & 	\cg{28.7}\std{0.5} & 	\cg{27.5}\std{0.3} & 	42.1\std{0.9} & 	60.2\std{0.5} & 	\cg{39.6}\std{0.7} & 	\cg{39.2}\std{0.7} \\
\bottomrule
\end{tabular}

\caption{
Object discovery on videos.
MOVi-C / E -- synthetic datasets; YTVIS-HQ -- real-world.
ARI -- mostly background segmentation accuracy; ARI\textsubscript{fg} -- foreground large objects; mBO -- best-matched segmentations, normalized by area; mIoU -- Hugarian-matched segmentations, normalized by area.
\texttt{\textcolor{red}{ECC}} and \texttt{\textcolor{green}{ICC}} are our explicit and implicit cycle consistency respectively.
Input resolution is 224$\times$224; DINO2 ViT-S/14 is employed for encoding; Using random seeds 42, 43 and 44.
}
\label{tab:objdiscov}
\end{table*}

\subsection{Video Object Discovery}
\label{sect:objdiscov}

\textbf{Metrics}.
Video object discovery performance intuitively reflects the quality of slots. We use standard unsupervised object segmentation metrics for the OCL setting:
Adjusted Rand Index (ARI)~\footnote{https://scikit-learn.org/stable/modules/generated/sklearn.metrics.adjusted\_rand\_score.html},
Foreground ARI (ARI\textsubscript{fg}),
Mean Best Overlap (mBO)~\citep{uijlings2013selectivesearch} and Mean Intersection over Union (mIoU)~\footnote{https://scikit-learn.org/stable/modules/generated/sklearn.metrics.jaccard\_score.html}.
We do not adopt Mean Average Precision (mAP) or similar metrics because in OCL there is no confidence-based precision-recall tradeoff.

\textbf{Datasets}.
We evaluate on multiple standard benchmarks.
MOVi-C and MOVi-E~\footnote{https://github.com/google-research/kubric/blob/main/challenges/movi}: synthetic videos featuring complex object dynamics (C, E) and camera movements (E).
YTVIS~\footnote{https://youtube-vos.org/dataset/vis} the High-Quality version~\footnote{https://github.com/SysCV/vmt?tab=readme-ov-file\#hq-ytvis-high-quality-video-instance-segmentation-dataset}: real-world YouTube videos with complex backgrounds, occlusions, motion / encoding blurs and textures.

\textbf{Baselines}.
We compare our method against recent representative video OCL methods.
VideoSAUR~\cite{zadaianchuk2024videosaur}: a classical method.
SlotContrast~\cite{manasyan2025slotcontrast}: SOTA in the year 2025.
RandSF.Q~\cite{zhao2025randsfq} and SmoothSA~\cite{zhao2025smoothsa}: the most recent SOTA methods that surpass SlotContrast further by a large margin.
To evaluate our implicit design, we incorporate our Implicit CC design into the two strongest baselines, denoted as RandSF.Q + \texttt{\textcolor{green}{ICC}} and SmoothSA + \texttt{\textcolor{green}{ICC}}.
We also evaluate ECC via RandSF.Q + \texttt{\textcolor{red}{ECC}} and SmoothSA + \texttt{\textcolor{red}{ECC}}.

\textbf{Codebase}.
Experiments are conducted using codebase \texttt{object-centric-bench}~\footnote{https://github.com/Genera1Z/RandSF.Q}~\footnote{https://github.com/Genera1Z/SmoothSA}
, which has reproduced many representative OCL methods with identical advanced data augmentation and training recipes~\citep{elsayed2022savipp}, ensuring fair, strong comparisons.  
It also provides model checkpoints and logs for all three standard random seeds, supporting reproducibility and efficient experimentation.

\textbf{Results}.
\Cref{tab:objdiscov} reports the results of all methods across the datasets and metrics.
The results show that ICC consistently improves ARI, ARI-FG, and mIoU across all datasets, with particularly strong gains in complex dynamic scenes. Further analysis reveals that ICC yields general performance improvements when integrated with RandSF.Q across all datasets and metrics, while on SmoothSA, slight degradations are observed on a few metrics.
Note that these results are upon two SOTA methods that already substantially outperform prior methods, making further improvements inherently difficult.
In contrast, the ECC objective on the basis methods always degenerates the performance.

\textbf{Efficiency}. As shown in \Cref{tab:overhead}, ICC introduces more computation overhead in both space and time than both the basis method and ECC.

\begin{table}[]
\centering\small

\begin{tabular}{ccccc}
\toprule
     & \multicolumn{2}{c}{training} & \multicolumn{2}{c}{evaluation} \\
\arrayrulecolor{gray}
\cmidrule(lr){2-3}\cmidrule(lr){4-5}
\arrayrulecolor{black}
per epoch; V100                 & GB & min & GB & min \\
\midrule
RandSF.Q @ MOVi-E               & 24.2 & 8.0 & 7.9 & 1.2 \\
+\texttt{\textcolor{red}{ECC}}  & 24.6 & 8.1 & 8.0 & 1.3 \\
+\texttt{\textcolor{green}{ICC}}& 24.7 & 8.4 & 8.4 & 1.3 \\
\bottomrule
\end{tabular}

\caption{
Computation overhead in space and time.
Spatial overhead is measured in peak VRAM consumption GB while temporal overhead is measured in time consumption minutes.
}
\label{tab:overhead}
\end{table}

\subsection{Video Object Recognition}
\label{sect:objrecogn}

\begin{table}[]
\small
\centering
\newcommand{\tss}[1]{\scalebox{0.8}{\texttt{#1}}}
\newcommand{\cg}[1]{\textcolor{green}{#1}}
\newcommand{\std}[1]{\scalebox{0.4}{±#1}}

\begin{tabular}{c@{}c@{}ccccc}
\toprule
\arrayrulecolor{gray}
&&& Top-1 & Top-3 & IoU & \#match \\
\cmidrule(lr){4-7}
&&& \multicolumn{4}{c}{YTVIS-HQ {\tiny \#slot=7}} \\
\arrayrulecolor{black}
\midrule
RandSF.Q    &+& MLP & 90.5\std{0.3} & 97.9\std{0.3} & 50.6\std{0.4} & 8979\std{123} \\
\quad + \textcolor{green}{\texttt{ICC}} & + & MLP & \cg{91.6}\std{0.2} & 97.7\std{0.3} & \cg{52.5}\std{0.5} & \cg{9233}\std{61} \\
\arrayrulecolor{gray}
\cmidrule(lr){1-3} \cmidrule(lr){4-7}
\arrayrulecolor{black}
SmoothSA & + & MLP & 90.4\std{0.2} & 97.6\std{0.1} & 42.6\std{1.4} & 8957\std{34} \\
\quad + \textcolor{green}{\texttt{ICC}} & + & MLP & \cg{91.5}\std{0.0} & \cg{97.9}\std{0.2} & \cg{47.0}\std{0.6} & \cg{9112}\std{48} \\
\arrayrulecolor{black}
\bottomrule
\end{tabular}

\caption{
Object recognition on videos.
Top-1 / Top-3: category classification accuracy; IoU: bounding box regression accuracy; \#match: number of matched objects.
By training a linear probe (MLP) on frozen slots from \Cref{tab:objdiscov}; Using random seeds 42, 43 and 44.
}
\label{tab:objrecogn}
\end{table}

Video object recognition performance directly measures the quality of slots. To verify if semantics beyond low-level texture are captured, we evaluate performance using \textbf{metrics} including Top-1 and Top-3 accuracy for category classification, IoU for bounding box regression, and the count of successfully matched objects (\#match). We conduct these evaluations on the real-world video \textbf{dataset} YTVIS-HQ. For \textbf{baselines}, we compare the vanilla counterparts of state-of-the-art methods RandSF.Q and SmoothSA against our ICC-integrated versions to demonstrate the improvements in latent space separation. All experiments are implemented within the same \textbf{codebase} as in object discovery, following the standard protocol: freezing the OCL model and training a lightweight 2-layer MLP to predict class labels and bounding boxes from the slot representations.

As shown in \Cref{tab:objrecogn} object recognition \textbf{results}, models trained with ICC outperform their vanilla counterparts in Top-1/Top-3 accuracy and box IoU. 
This indicates that our consensus objective forces the slots to retain more discriminative identity features rather than just low-level texture information, thereby facilitating better separation of object categories.

\subsection{Ablation Study}
\label{sect:ablat}

\textbf{Does the performance gain come from the extra backward reconstruction?} -- No. 
We design a new experiment item, Non-Chain Reconstruction (NCR), which replaces the chained initialization in \Cref{eq:bw_initializ_i} with the default initialization similar to \Cref{eq:initaliz}. Namely, dependence between the forward and backward streams are removed; they are just two parallel streams in inverse direction. In this setting, there is still the extra backward reconstruction. But as shown in \Cref{tab:source_of_gains}, NCR shows no consistent superiority.

\textbf{Is that ECC performs worse than ICC due to its naive implementation?} -- No.
We design a stronger ECC baseline, Hungarian ECC. We enforce the ECC loss on Hungarian matched slot pairs, rather than on slot pairs that have identical indexes in two sets of slots, as what is conducted in \Cref{eq:bw_loss_explicit} originally. The match metric is cosine similarity. This experiment item can handle the identity switch along time, thus can perform better. As shown in \Cref{tab:source_of_gains}, although the Hungarian ECC is a bit better but still much worse than ICC.

\begin{table}[]
\centering\small
\newcommand{\tss}[1]{\scalebox{0.8}{\texttt{#1}}}
\newcommand{\cg}[1]{\textcolor{green}{#1}}
\newcommand{\ch}[1]{\textcolor{red}{#1}}
\newcommand{\std}[1]{\scalebox{0.4}{±#1}}

\begin{tabular}{lcccc}
\toprule
@MOVi-C & ARI & ARI\textsubscript{fg} & mBO & mIoU \\
\midrule
RandSF.Q
& 65.4\std{10.7} & 67.4\std{2.1} & 29.2\std{3.8} & 26.8\std{3.7} \\
\arrayrulecolor{gray}
\cmidrule(lr){1-1} \cmidrule(lr){2-5}
+ \texttt{\textcolor{red}{ECC}}
& \ch{53.8}\std{3.4} & \ch{46.6}\std{3.7} & \ch{20.5}\std{1.1} & \ch{17.6}\std{1.7} \\
+ Hungarian \texttt{\textcolor{red}{ECC}}
& \ch{59.4}\std{2.7} & \ch{47.2}\std{5.3} & \ch{22.4}\std{2.5} & \ch{18.3}\std{2.8} \\
\cmidrule(lr){1-1} \cmidrule(lr){2-5}
+ Non-Chain Recon.
& 68.3\std{1.5} & 65.1\std{8.8} & 29.1\std{1.4} & 27.4\std{1.5} \\
\cmidrule(lr){1-1} \cmidrule(lr){2-5}
+ \texttt{\textcolor{green}{ICC}}
& \cg{73.2}\std{0.7} & \cg{67.4}\std{1.0} & \cg{32.9}\std{0.4} & \cg{30.3}\std{0.3} \\
\arrayrulecolor{black}
\bottomrule
\end{tabular}

\caption{
Isolating the source of performance gains. Hungarian ECC isolates permutation drift from decomposition divergence; Non-Chain Reconstruction represents doubled reconstruction without a temporal chain as in ICC.
}
\label{tab:source_of_gains}
\end{table}

\section{Dissection: Mechanism of Implicit Consensus}
\label{sect:dissect}

\subsection{Quantifying Feature Collapse vs. Diversity}
\label{sect:dissect_collapse}

We argue in \Cref{sect:cc_explicit} that ECC hard alignment forces slots towards a mean-collapsed state. To quantify this, we measure the \textbf{slot variance}, defined as the average variance of slot features across the temporal dimension for a tracked object, and the \textbf{slot diversity}, measuring the cosine distance between slots within a frame.

As shown in Table~\ref{tab:collapse_analysis}, applying Explicit CC to RandSF.Q results in a sharp drop in slot diversity, i.e.,  0.934 $\to$ 0.904, confirming that the slots become ``averaged'' and lose discriminative identity. In contrast, our Implicit CC maintains high diversity 0.300 comparable to the baseline 0.298, without worsening reconstruction error. This proves that ICC aggregates slots through time without sacrificing representation distinctiveness.

\begin{table}[]
\centering
\small
\newcommand{\tss}[1]{\scalebox{0.8}{\texttt{#1}}}
\newcommand{\cg}[1]{\textcolor{green}{#1}}
\newcommand{\ch}[1]{\textcolor{red}{#1}}
\newcommand{\std}[1]{\scalebox{0.4}{±#1}}

\begin{tabular}{ccccc}
\toprule
@YTVIS-HQ & \multicolumn{2}{c}{slot} & recon. \\
\arrayrulecolor{gray}
\cmidrule(lr){2-3} \cmidrule(lr){4-4}
\arrayrulecolor{black}
\textcolor{gray}{$\times$100} & diversity\textuparrow & variance\textuparrow & error\textdownarrow \\
\midrule
RandSF.Q            & 93.4\std{0.9} & 29.8\std{1.5} & 51.5\std{3.1} \\
\quad + \texttt{\ch{ECC}}& \ch{90.4}\std{1.2} & \ch{19.5}\std{1.3} & \ch{67.3}\std{6.4} & \textit{\ch{collapse}}\\
\quad + \texttt{\cg{ICC}}& 91.6\std{0.9} & 30.0\std{2.4} & 53.7\std{5.9} \\
\arrayrulecolor{black}
\bottomrule
\end{tabular}

\caption{
Quantifying representation collapse.
\texttt{\textcolor{red}{ECC}} drastically reduces slot diversity (feature \textcolor{red}{collapse}), whereas \texttt{\textcolor{green}{ICC}} maintains spatial diversity and temporal variance without worsening reconstruction error.
}
\label{tab:collapse_analysis}
\end{table}

\begin{figure}[]
\small
\centering
\includegraphics[width=0.5\linewidth]{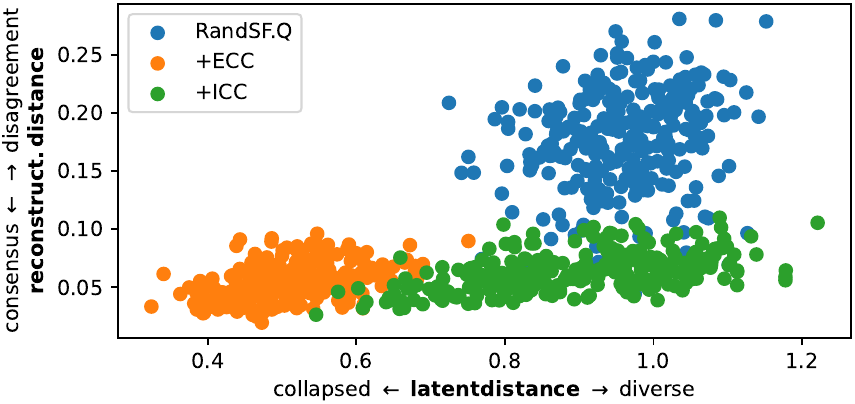}
\caption{
Manifold Alignment Analysis.
Each dot represents a video frame. \texttt{\textcolor{green}{ICC}} achieves high consensus on reconstruction (low Y-axis) despite allowing slots to diverge in the latent space (high X-axis), validating that we align on the reconstruction manifold.
}
\label{fig:manifold}
\end{figure}

\subsection{The Manifold Alignment Hypothesis}
\label{sect:dissect_manifold}

We argue in \Cref{sect:cc_implicit} that ICC align streams on the observation manifold, achieving consensus in the observation / reconstruction space while maintaining diversity in the latent / slot space.
We analyze the relationship between them using \textbf{latent distance}, measuring slot distance between forward and backward streams, and \textbf{reconstruction distance}, measuring the reconstruction error between forward and backward streams. As there is no backward stream in the baseline, we run the baseline model on videos played inversely; Similarly as ECC has no backward stream reconstruction, we reuse the decoder to decode the backward slots into quasi-backward reconstruction.

\Cref{fig:manifold} plots these two metrics for YTVIS-HQ videos. 
ECC results cluster in the bottom-left: low reconstruction disagreement but at the cost of latent collapse.
Baseline RandSF.Q is scattered: high latent distance, high reconstruction disagreement.
ICC forms a unique cluster in the \textit{bottom-right}: high latent diversity and better reconstruction consensus.
This empirically proves that our method successfully decouples latent similarity from semantic consistency, allowing the model to navigate the solution space flexibly.

\subsection{Visualizing Decomposition Divergence}
\label{sect:dissect_visualiz}

A key motivation for our method is that the Forward and Backward streams may generate distinct but equally valid segmentations. ECC penalizes this valid ambiguity.

In Figure~\ref{fig:diverge}, we visualize the attention masks from the Forward and Backward streams at the timestep $t$. 
(\textit{middle}) \texttt{\textcolor{red}{ECC}} forces the masks to be identical. Since the streams disagree on the ``correct'' decomposition, the model outputs blurry, uncertain masks, failing to capture the object.
(\textit{right}) For \texttt{\textcolor{green}{ICC}}, both streams clearly separate the car, land, trees and sky, which is crucial. But specifically for the car, these streams have different decompositions schema.
By aligning on the \textit{reconstruction} rather than \textit{slots}, ICC allows such semantic flexibility.

\section{Conclusion}

We demonstrate that applying explicit cycle consistency to OCL is ill-posed, as rigid latent alignment conflicts with the stochastic nature of scene decomposition, leading to feature collapse. To handle this, we propose Implicit Cycle Consistency (ICC), which aligns forward-backward streams on the reconstruction manifold. This approach enforces temporal coherence while allowing slot representations to diverge, successfully reconciling object stability with valid decomposition ambiguity.
This work sets a good starting point for more techniques from unsupervised MOT to be explored in video OCL.

\bibliography{egbib}

\end{document}